\documentclass[10pt,twocolumn,letterpaper]{article}

\usepackage{3dv}
\usepackage{times}
\usepackage{epsfig}
\usepackage{graphicx}
\usepackage{amsmath}
\usepackage{amssymb}

\usepackage[pagebackref=true,breaklinks=true,colorlinks,bookmarks=false]{hyperref}
\usepackage{enumitem}

\usepackage{bm}
\usepackage{caption} 
\usepackage[capitalize]{cleveref}
\usepackage{booktabs}

\usepackage{algorithm}
\usepackage{algorithmicx}
\usepackage{algpseudocode}
\usepackage{amsthm}
\usepackage{mathrsfs}

\crefname{section}{Sec.}{Secs.}
\Crefname{section}{Section}{Sections}
\Crefname{table}{Table}{Tables}
\crefname{table}{Tab.}{Tabs.}

\threedvfinalcopy 

\def\threedvPaperID{138} 

\begin{document}
	
\title{Dual-Space NeRF: \\Learning Animatable Avatars and Scene Lighting in Separate Spaces}

\author{
	Yihao Zhi\textsuperscript{\rm 1}\thanks{Equal contribution.} \quad
	Shenhan Qian\textsuperscript{\rm 1}\footnotemark[1] \quad 
	Xinhao Yan\textsuperscript{\rm 1}\footnotemark[1] \quad 
	Shenghua Gao\textsuperscript{\rm 1,2,3}\thanks{Corresponding author.}\\
	{\tt\small \{zhiyh, qianshh, yanxh, gaoshh\}@shanghaitech.edu.cn}
	\\ \\
	\textsuperscript{\rm 1}ShanghaiTech University \\
	\textsuperscript{\rm 2}Shanghai Engineering Research Center of Intelligent Vision and Imaging \\
	\textsuperscript{\rm 3}Shanghai Engineering Research Center of Energy Efficient and Custom AI IC
}

\maketitle
\thispagestyle{empty}

\begin{abstract}
	Modeling the human body in a canonical space is a common practice for capturing and animation. But when involving the neural radiance field (NeRF), learning a static NeRF in the canonical space is not enough because the lighting of the body changes when the person moves even though the scene lighting is constant. Previous methods alleviate the inconsistency of lighting by learning a per-frame embedding, but this operation does not generalize to unseen poses. Given that the lighting condition is static in the world space while the human body is consistent in the canonical space, we propose a dual-space NeRF that models the scene lighting and the human body with two MLPs in two separate spaces. To bridge these two spaces, previous methods mostly rely on the linear blend skinning (LBS) algorithm. However, the blending weights for LBS of a dynamic neural field are intractable and thus are usually memorized with another MLP, which does not generalize to novel poses. Although it is possible to borrow the blending weights of a parametric mesh such as SMPL, the interpolation operation introduces more artifacts. In this paper, we propose to use the barycentric mapping, which can directly generalize to unseen poses and surprisingly achieves superior results than LBS with neural blending weights. Quantitative and qualitative results on the Human3.6M and the ZJU-MoCap datasets show the effectiveness of our method. Our code is available at: \href{https://github.com/zyhbili/Dual-Space-NeRF}{https://github.com/zyhbili/Dual-Space-NeRF}.
	
\end{abstract}

\section{Introduction}
\label{sec:intro}

Human body reconstruction and rendering have long been an active research topic. Multi-view videos are especially suitable for this task because they record not only the appearance but also the movements and deformation of a person. Classic reconstruction and rendering techniques show limited image realism due to the complexity of decoupling the geometry, material and lighting from images. However, the recently proposed neural radiance field (NeRF) \cite{mildenhall2020nerf} proves it possible to represent a static scene with an MLP without explicitly modeling the above factors. Several recent works \cite{peng2021neural,liu2021neural,peng2021animatable,2021narf,su2021anerf,weng2022humannerf,xu2021h,kwon2021neural} have adapted NeRF onto human body reconstruction and animation, but challenges remain in the following aspects:

\begin{figure}[t]
	\centering
	\includegraphics[width=0.99\linewidth]{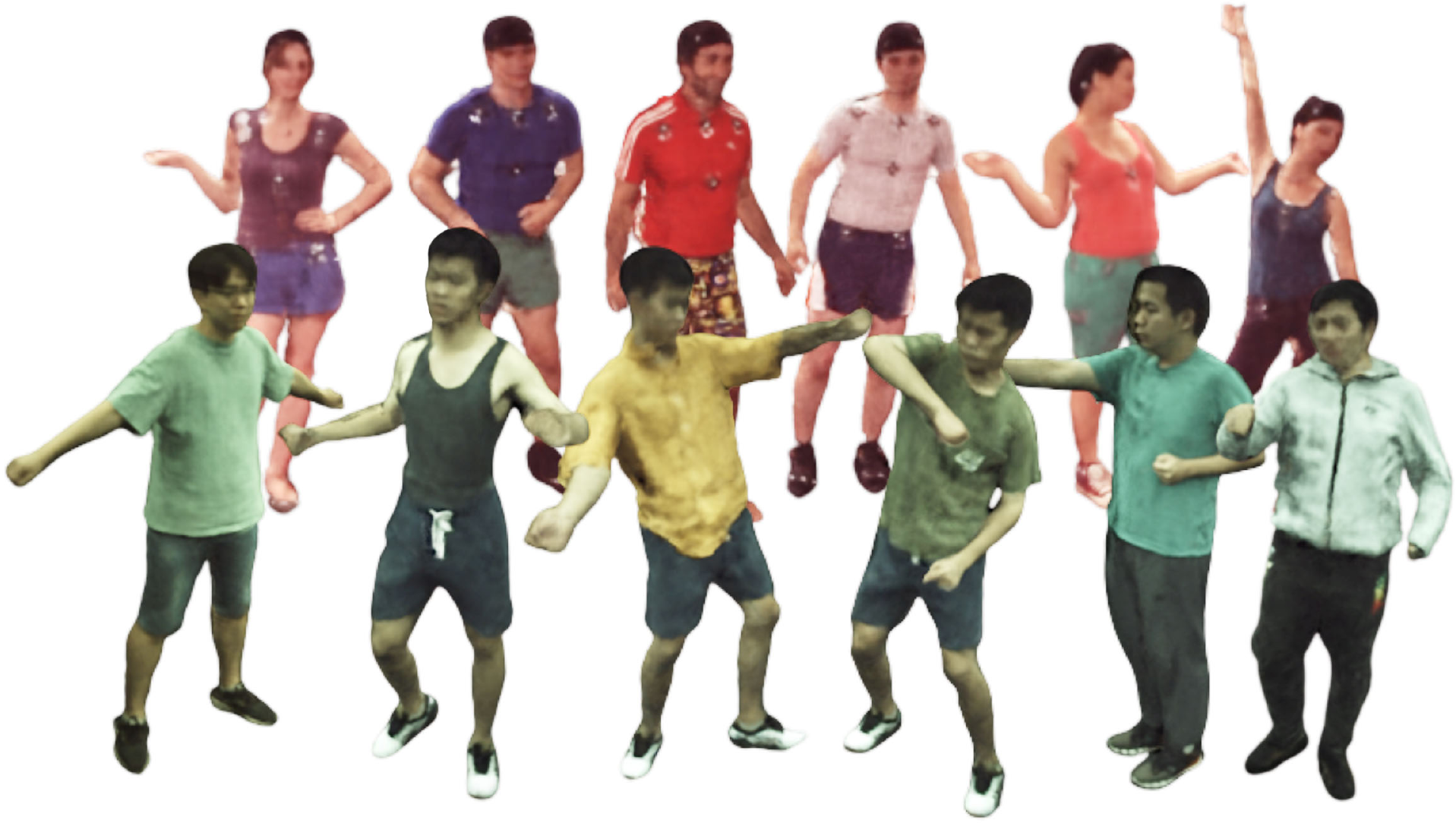}
	\caption{Our method learns a human body with a neural radiance field \cite{mildenhall2020nerf} and animates it in arbitrary poses with no need for fine-tuning or additional input. Since we model the subject in the canonical space and the lighting in the world space, we name the method ``Dual-Space NeRF".}
	\label{fig:teaser}
\end{figure}

Recent methods that model the human body with NeRF~\cite{mildenhall2020nerf} mostly learn the human body in a canonical space, but the lighting inconsistency in the canonical space lacks exploration \cite{peng2021neural, liu2021neural, peng2021animatable, 2021narf, su2021anerf}. When learning a NeRF in the canonical space, we assume the appearance of a person is consistent across varied body poses. But the fact is when a person moves, a point in the canonical space comes to a different place in the world space, resulting in a change of the lighting condition. This means that merely modeling the appearance of a person in the canonical space is not enough. Therefore, we propose to model the scene lighting with another MLP in the world space, where the scene lighting is assumed to be static. The lighting MLP takes in a point position, a normal vector, and a view direction in the world space, and outputs a lightness coefficient that adjusts the brightness of the point.

Different from NeRF \cite{mildenhall2020nerf} that only depends on the point position and the viewing direction, our lighting MLP also takes in a normal vector, still due to the complexity of dynamic scenes. For a static scene, the surface normal can be uniquely determined by the point position, while for our setting, the surface normal may change when the subject moves. Unlike IDR \cite{yariv2020multiview} that models the lighting condition and the appearance of an object with one appearance MLP, our method learns the two factors with separate MLPs in different spaces to realize correct lighting under unseen poses. Also, our lighting MLP only predicts a scalar coefficient to rescale the color from the body MLP instead of directly predicting a color to prevent overfitting.

To bridge the world space and the canonical space, the key is building pixel-level correspondences across views and frames. A stream of methods bind points on 3D skeletons \cite{2021narf,su2021anerf} based on the assumption of local rigidity. Another stream of methods incorporates geometric priors characterized by anchoring points onto SMPL \cite{SMPL:2015}, a parametric human body model. Neural Body \cite{peng2021neural} binds features onto SMPL's vertices and diffuses them into the space before volumetric rendering. It produces realistic novel views of the training sequence but degrades on novel poses. Animatable NeRF \cite{peng2021animatable} resolves novel-pose synthesis by mapping observed points into a canonical space with inverse linear blend skinning (LBS). Since the LBS weight of a spatial point varies for different poses, Animatable NeRF \cite{peng2021animatable} learns a neural blending weight network conditioned on the pose, which requires additional training for novel poses.

To avoid learning the volatile LBS weights, we seek a pose-independent local position representation that generalizes to novel poses easily. Specifically, we propose a barycentric mapping (BM) as follows. For a point in the space, we first project it onto its closest face on the fitted SMPL mesh. Then we describe this point by the barycentric coordinates of its projected point and its signed height from the face. Finally, its corresponding point in the canonical space is uniquely determined. Note that NPMs \cite{palafox2021npms} leverages a similar barycentric mapping to obtain pseudo ground truths to train a unidirectional deformation field. While in our method, we extend BM to support vector transformation and use it bidirectionally, transforming a point position from the world space to the canonical space and warping a surface normal from the canonical space to the world space. By this barycentric mapping, the body MLP and the lighting MLP are bridged. BM is parameter-free, enabling pose generalization without additional input or network fine-tuning. Though the parameter-free method seems to have inferior expressiveness, our experiments show its comparable ability on two datasets and clear advantages under challenging poses. It is flexible since it anchors a point on the edge vectors of a face, allowing local deformation along with the face. It also avoids the artifacts of inverse LBS caused by blending weighting interpolation (\cref{fig:ablation_mapping}).

Another challenge of this task is the existence of random variations such as clothing wrinkles. These variations are neither fully determined by the body pose nor consistent across frames, making it harder to learn a stable canonical radiance field. Neural Actor \cite{liu2021neural} uses the ground-truth texture map to relieve the ambiguity in training images but requires a separate image generation network to infer texture maps from normal maps for testing. Motivated by SCANimate \cite{Saito:CVPR:2021}, to model the pose-dependent deformation, our model is conditioned on the pose parameters. Besides, per-frame latent embeddings are used to capture the random variations. According to our experiments, the union of these components is sufficient to represent vivid deformations along with pose changes with no need for an extra deformation network. 

Our contributions can be summarized as follow:
\begin{itemize}[itemsep=2pt,topsep=0pt,parsep=0pt]
	\item To ensure the lighting correctness under unseen poses, we propose dual-space NeRF, which models the static scene lighting in the world space and the human body in the canonical space.
	\item We propose to use the barycentric mapping to build correspondences between two spaces and validate its comparable expressiveness and superior generalization ability under extreme poses.
	\item We show the effectiveness and interpretability of our method with quantitative and qualitative results on the Human3.6M \cite{ionescu2013human3} and the ZJU-MoCap \cite{peng2021neural} datasets.
\end{itemize}

\begin{figure*}[t]
	\centering
	\includegraphics[width=1\linewidth]{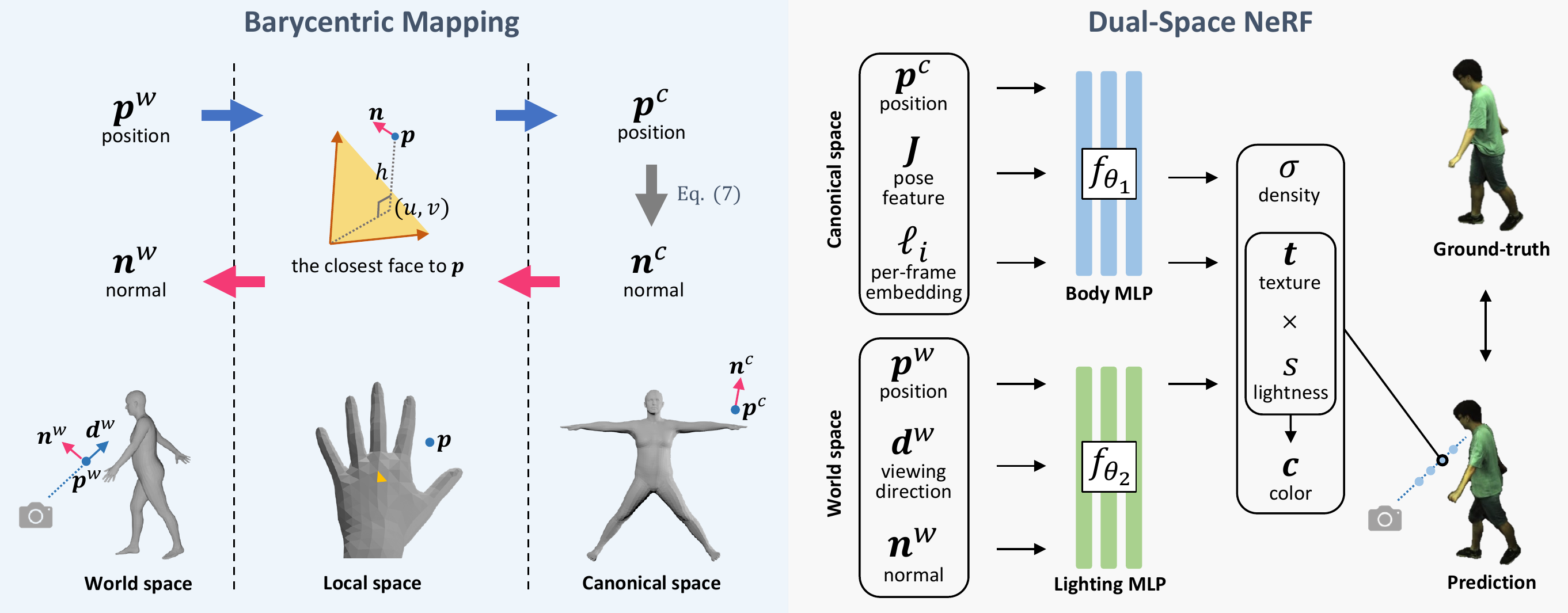}
	\caption{The pipeline of our method. Given a point $\bm{p}^{w}$ in the world space, we use the barycentric mapping to warp a point $\bm{p}^{w}$ from the world space into the canonical space to query the body properties. In the canonical space, we compute the surface normal $\bm{n}^{c}$ and warp it back to obtain the normal vector $\bm{n}^{w}$ in the world space. We learn a Body MLP to model a human body in the canonical space and a Lighting MLP to capture the lighting condition in the world space. Finally, we render an image with volumetric rendering.}
	\label{fig:pipeline}
\end{figure*}

\section{Related Work}
\label{sec:related}
\textbf{3D human reconstruction.}
As a popular 3D articulated human model, SMPL \cite{SMPL:2015} learns a template mesh from 3D scans and deforms the mesh with the linear blend skinning (LBS) algorithm. The mesh-based 3D human model is easy to manipulate but limited to a fixed topology. Therefore, neural implicit functions are adopted to model a 3D avatar that can be animated by SMPL \cite{Saito:CVPR:2021,chen2021snarf,deng2020nasa,LEAP:CVPR:21}. 
LEAP \cite{LEAP:CVPR:21}, SCANimate \etal \cite{Saito:CVPR:2021}, and SNARF \cite{chen2021snarf} learn the human body in a canonical space and articulate the body with neural blending weights. NASA \cite{deng2020nasa} is a part-based method that binds an implicit function on each bone. These methods use 3D data as the input and only account the shape of the human body.

\textbf{Dynamic neural radiance field.}
Neural radiance fields (NeRF) \cite{mildenhall2020nerf} can synthesize photorealistic images from arbitrary views with no need for 3D data. However, the vanilla NeRF is designed for only static scenes. The challenge for NeRF to model dynamic scenes lies in building correspondences across the timeline. Recent methods define a transformation field that warps an observed point into a canonical space \cite{pumarola2021d,tretschk2021non,park2021nerfies,park2021hypernerf,guo2021ad}.
HumanNeRF \cite{weng2022humannerf} records the variation of a scene by learning a motion field from monocular video. Kwon \etal \cite{kwon2021neural} resort to a temporal transformer to aggregate skeletal, temporal, and spatial features. These methods exhibit strong ability of replaying events in novel views but are incapable of generating new contents such as animating a human body under unseen poses.

\textbf{Human body animation with NeRF.}
To animate a NeRF of a human body, a straightforward solution is incorporating 3D human priors. Peng \etal \cite{peng2021neural} use a set of latent code to encode the local geometry and appearance of the human body and bind them onto SMPL\cite{SMPL:2015} vertices. Liu \etal \cite{liu2021neural} introduce Neural Actor, animating NeRF with blending weights sampled from the nearest vertex of SMPL. Additionally, an image translation network is used to infer texture maps to provide residual deformations and appearance details for novel poses. AniNeRF \cite{peng2021animatable} learns a neural blending weight field to learn the LBS weights for each particular pose. Since the blending weight field varies with poses, AniNeRF relies on a per-frame latent vector as a condition for training poses and requires fine-tuning for novel poses. Xu \etal \cite{xu2021h} learn NeRF upon imGHUM \cite{alldieck2021imghum}, a statistical human body model represented by neural implicit functions.

\section{A Revisit of NeRF}
NeRF \cite{mildenhall2020nerf} represents a scene by density $\sigma$ and color $\bm{c}$ at each spatial point $\bm{p}$. To render an image in an arbitrary view, $\sigma$ and $\bm{c}$ are accumulated along viewing rays. 
Formally, we denote a viewing ray emitted from the optical center of a camera through a given pixel on the image plane by $\bm{r}(m) = \bm{o} + m \bm{d}$, then an approximation of the pixel color is
\begin{equation}
	\hat{\mathbf{C}}(\bm{r}) = \mathcal{R}(\bm{r}, \bm{c}, \sigma) = \sum^{K}_{k=1} T(m_k) \alpha(\sigma(m_k) \delta_k) \bm{c}(m_k),
	\label{eq:volumetric_render}
\end{equation}
where $\mathcal{R}(\bm{r}, \bm{c}, \sigma)$ is the volumetric rendering of the color $\bm{c}$ with the density $\sigma$; $\left\{m_k\right\}^K_{k=1}$ is a set of discretely sampled points between the near and the far plane of the camera; $\delta_k= m_{k+1} - m_k$ is the distance between the current sampling point and the next one; $T(m_k) = \exp \left(-\sum^{k-1}_{k'=1} {\sigma(m_{k'}) \delta_{k'}} \right)$, and $\alpha(x)=1-\exp(-x)$. NeRF learns a radiance field with an MLP in the form of
\begin{equation}
	\left[\sigma(m), \bm{c}(m)\right] = f_{\theta}(\gamma_{\bm{p}}(\bm{r}(m)), \gamma_{\bm{d}}(\bm{d})), \label{eq:f2}
\end{equation}
where $\theta$ is the model parameter, $\gamma_{\bm{p}}(\cdot)$ and $\gamma_{\bm{d}}(\cdot)$ are fixed positional encoding functions for positions and directions. 

The network parameters are optimized by the loss
\begin{equation}
	\mathcal{L} = \frac{1}{NM} \sum^{N}_{i=1} \sum^{M}_{j=1} \left\Vert \mathbf{C}(\bm{r}_{ij}) - \hat{\mathbf{C}}(\bm{r}_{ij}) \right\Vert _2^2,
\end{equation}
where $N$ is the number of images, $M$ is the number of rays in each image, and $\bm{r}_{ij}$ is the $j^{\text{th}}$ ray in the $i^{\text{th}}$ image.

\section{Method}
Our method learns to reconstruct a person from synchronized multi-view video frames and animates the subject with novel poses. This is achieved by learning a canonical neural radiance field \cite{mildenhall2020nerf} of a human body in X-pose. This canonical radiance field is anchored to SMPL \cite{SMPL:2015} so that we can animate the radiance field by manipulating SMPL. 
In \cref{fig:pipeline}, we show the pipeline of our method with barycentric mapping (\cref{sec:barycentric}) and dual-space NeRF (\cref{sec:dualspace_nerf}). The dual-space NeRF includes two networks: a Body MLP (\cref{sec:bodymlp}) to model a human body in the canonical space and a Lighting MLP (\cref{sec:lightingmlp}) to capture the location-dependent lighting in the world space. And the barycentric mapping bridges the canonical space and the world space.

\subsection{Barycentric Mapping}
\label{sec:barycentric}
Considering the sparsity of views and the ambiguity caused by smooth regions, it is tough to learn robust correspondences across frames purely with images. Therefore, we adopt SMPL \cite{SMPL:2015} as a geometric prior of the human body by anchoring spatial points on the faces of SMPL.

\subsubsection{Position mapping}
\label{sec:position_mapping}
\begin{figure}[h!]
	\centering
	\includegraphics[width=0.8\linewidth]{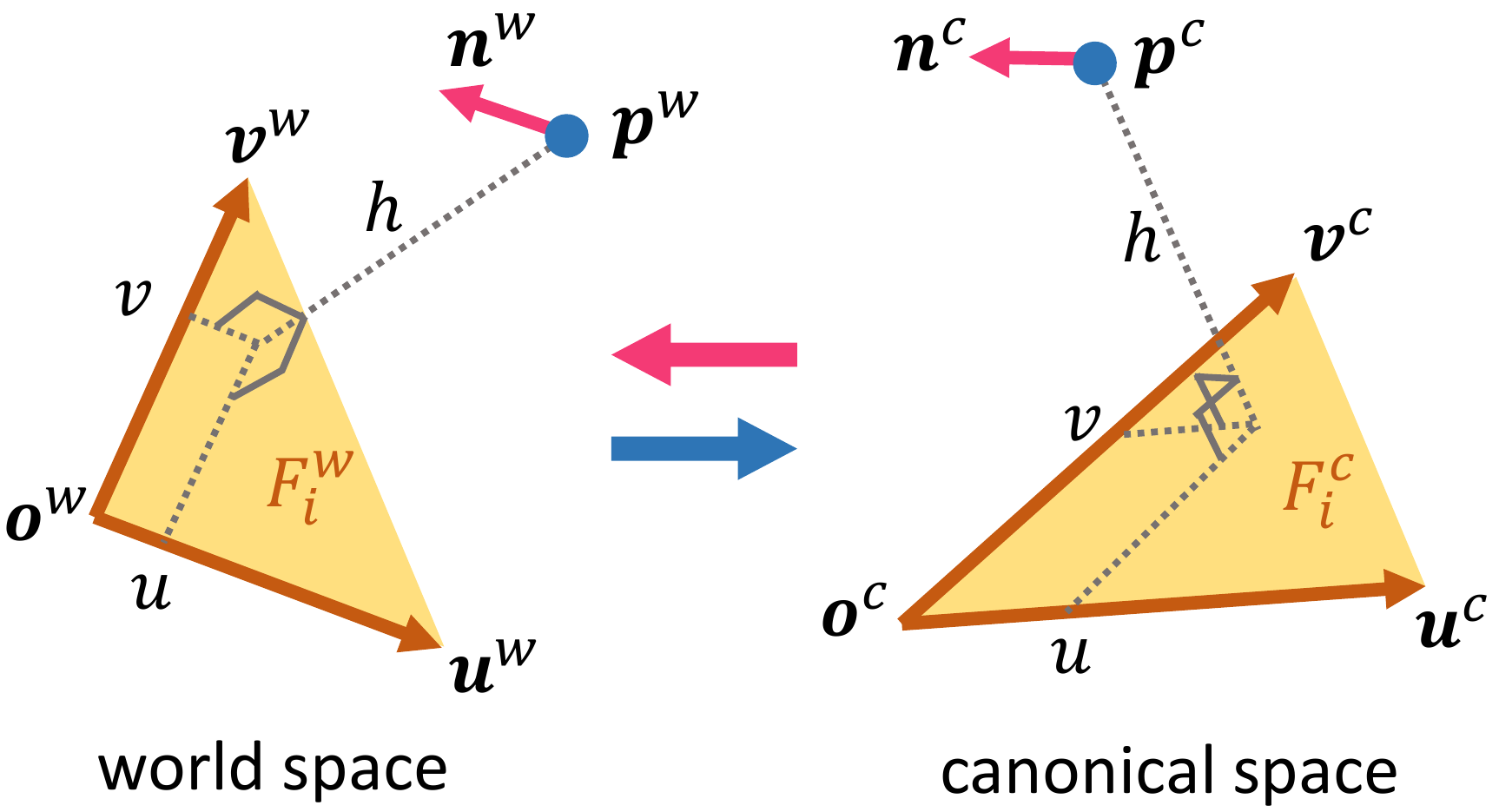}
	\caption{The barycentric mapping of positions and directions. For a point $\bm{p}^w$ in the world space, we first find its closest face $F_i^w$, whose corresponding face in the canonical space is $F_i^c$. Then, we represent $\bm{p}^w$ as the barycentric coordinate $(u, v)$ of its projection on $F_i^w$ and its signed distance $h$ from $F_i^w$. Finally, we compute the counterpart  of $\bm{p}^w$ in the canonical space, \ie, $\bm{p}^c$, with the same local representation upon $F_i^c$. The barycentric mapping can also be used to transform direction vectors.}
	\label{fig:barycentricmap}
\end{figure}

For a point $\bm{p}^w$ in the world space (\cref{fig:barycentricmap}), we first determine its closest face $F_i^w$ by measuring its distances to the mean of vertex positions of each face. Then, we set up a local description of the point $\bm{p}^w$ by $(u,v,h)$, where $(u,v)$ is the barycentric coordinate of the projection of $\bm{p}^w$ on the face $F_i^w$, and $h$ is the signed distance from $F_i^w$. Based on the corresponding face of $F_i^w$ in the canonical space, \ie, $F_i^c$, we can compute the corresponding point of $\bm{p}^w$ as:
\begin{equation}
	\bm{p}^c = \bm{o}^{c} + u\bm{u}^{c} + v\bm{v}^{c} + h  \frac{\bm{u}^{c} \times \bm{v}^{c}}{\|\bm{u}^{c} \times \bm{v}^{c}\|},
\end{equation} 
where $\bm{o}^{c}$ is the first vertex of the face $F_i^c$, $\bm{u}^{c}$ and $\bm{v}^{c}$ are two edge vectors of $F_i^c$ starting from the vertex $\bm{o}^{c}$. Note that the mapping can be conducted in an inverse direction.

\subsubsection{Direction mapping}
\label{sec:direction_mapping}
Based on the position mapping, we can bridge direction vectors between spaces, also in a differentiable manner. For the example in \cref{fig:barycentricmap}, we first represent the direction vector $\bm{n}^c$ by its starting point $\bm{p}^c$ and its ending point $\bm{p}^{c}_e = \bm{p}^c + \bm{n}^c$. Then we apply the position mapping described above to get the corresponding positions in the world space, \ie, $\bm{p}^{w}$ and $\bm{p}^{w}_e$. Finally, the warped direction vector in the world space can be obtained by:
\begin{equation}
	\bm{n}^w = \frac{\bm{p}^w_e - \bm{p}^w}{\| \bm{p}^w_e - \bm{p}^w \|}.
\end{equation}

\subsection{Dual-Space NeRF}
\label{sec:dualspace_nerf}
NeRF \cite{mildenhall2020nerf} astonishes the community for its high rendering realism, especially for the view-dependent visual effects. Most importantly, NeRF is formulated as a function of merely a point position $\bm{p}$ and a viewing direction $\bm{d}$. Physically, the shading of a point also depends on the lighting condition and the surface normal, but NeRF omits them because they can be fully determined by the point position for a static scene.

However, for animatable human reconstruction and animation, the lighting condition is static only in the world space while the body shape and the surface normal is consistent only in the canonical space. Therefore, we have to learn a Body MLP and a Lighting MLP in separate spaces.

\subsubsection{Body MLP}
\label{sec:bodymlp}
Given a point position $\bm{p}^w$ in the world space (also called the observed space), we use the barycentric mapping (\cref{sec:position_mapping}) to obtain its corresponding point in the canonical space, \ie, $\bm{p}^c$, which is the  main input of the Body MLP. Motivated by SCANimate \cite{Saito:CVPR:2021}, we encode the quaternion matrix of the joints of SMPL with a tiny MLP and get the pose feature $\bm{J}$. To prevent artifacts and blur in the results, we also learn a latent embedding $\bm{\ell}_i \in \mathbb{R}^8$ for each video frame $i$ to model the random variations that cannot be fully determined by the body pose. 

Formally, we feed the canonical point position $\bm{p}^c$, pose features $\bm{J}$, and latent embedding $\bm{\ell}_i$ into the Body MLP to predict density $\sigma$ and texture $\bm{t} \in \mathbb{R}^{3}$. Here, \cref{eq:f2} is re-formulated as:
\begin{equation}
	\left[\sigma, \bm{t}\right] = f_{\theta_1}(\gamma_{\bm{p}}(\bm{p}^c), \bm{J} , \bm{\ell}_i).\label{eq:f4}
\end{equation}
For the Body MLP, density $\sigma$ models the static shape and texture $\bm{t}$ models the true color of a human body in X-pose, both independent of the viewing direction. Moreover, the surface normal at the position $\bm{p}^c$ can be obtained by the normalized gradient of density $\sigma$ with respect to $\bm{p}^c$ \cite{nerv2021,boss2021nerd}:
\begin{equation}
	\bm{n}^{c} = - \frac{\nabla \sigma}{\|\nabla \sigma\|}. \label{eq:normal}
\end{equation}

\subsubsection{Lighting MLP}
\label{sec:lightingmlp}
To illustrate the necessity of a separate Lighting MLP, we consider a point $\bm{p}^c$ on a hand of a subject in the canonical space. When the subject waves the hand, the corresponding location of $\bm{p}^c$ in the world space moves, resulting in a change of the lighting condition. Therefore, from the perspective of the point $\bm{p}^c$, the lighting condition varies with the body pose. Although the per-frame (or per-pose) lighting embeddings used by previous methods \cite{peng2021animatable,park2021nerfies} do help relieve the inconsistency of lighting in the training frames, they cannot generalize to novel poses. In contrast, we learn a Lighting MLP in the world space, making the lighting condition independent of the body pose.

Concretely, we use the Lighting MLP to predict a lightness coefficient
\begin{equation}
	s = f_{\theta_2}(\bm{p}^{w}, \bm{d}^w, \bm{n}^{w}),
\end{equation}
where $\bm{p}^{w}$ is the point position, $\bm{d}^w$ is the viewing direction, and $\bm{n}^w$ is the surface normal, all in the world space. During ray casting and point sampling, $\bm{p}^w$ and $\bm{d}^w$ are directly available while the surface normal $\bm{n}^w$ is not. To obtain $\bm{n}^w$, we first use the barycentric mapping (\cref{sec:position_mapping}) to find the corresponding point of $\bm{p}^w$ in the canonical space, \ie, $\bm{p}^c$, then compute the surface normal of $\bm{p}^c$,\ie, $\bm{n}^c$, finally map $\bm{n}^c$ back to the world space also with the barycentric mapping (\cref{sec:direction_mapping}).
The lightness coefficient $s$ is meant to scale the lightness of the texture $\bm{t}$ for shading with the color
\begin{equation}
	\bm{c} = s \bm{t},
\end{equation}
and the final result is obtained by volumetric rendering with the density $\sigma$ and the color $\bm{c}$. Here, we model the lighting condition with simply a lightness coefficient instead of a color vector or more complex models because the task is highly under-constrained, and suppressing the expressiveness of the Lighting MLP helps prevent overfitting (\cref{fig:ablation_lightingmlp_design}).

\subsection{Implementation Details}
We apply the neutral SMPL \cite{SMPL:2015} model for body mesh fitting.
Personalized shape parameters are used for each subject. The X-pose is chosen as the canonical pose. 
Our network consists of 2 parts: the Body MLP is an 8-layers MLP with a shortcut connected to the fifth layer, and the Lighting MLP is a 4-layers MLP. We use a 3-layer MLP to learn the pose feature $\bm{J}$. Hidden layers are activated by ReLU. The per-frame embeddings are initialized with Gaussian distribution ($\mathcal{N}(0, 1)$). 
We use the Adam optimizer to train our network for 200 epochs. We set the learning rate to 0.0005 and exponentially downscale it until the last epoch to 10 times lower. Weight decay is not used. 
We use a batch size of 1 with 5000 rays per batch. We sample 64 points on each ray. 
To accelerate training and inference, we abandon the coarse-to-fine strategy \cite{mildenhall2020nerf,peng2021neural,peng2021animatable}. Instead, we adopt the geometry-guided ray marching \cite{liu2021neural}, which produces a tighter bound. 
Since we have instance-level human-parsing masks, we ensure that 5 percent of rays are sampled around the face in each iteration, and the other rays are randomly sampled in the 2D bounding box. 
All experiments are conducted on a GeForce RTX 2080 Ti GPU and take around two days to converge.

\section{Experiments}
\label{sec:exp}

\subsection{Settings}
\begin{figure*}[t]
	\centering
	\includegraphics[width=1\linewidth]{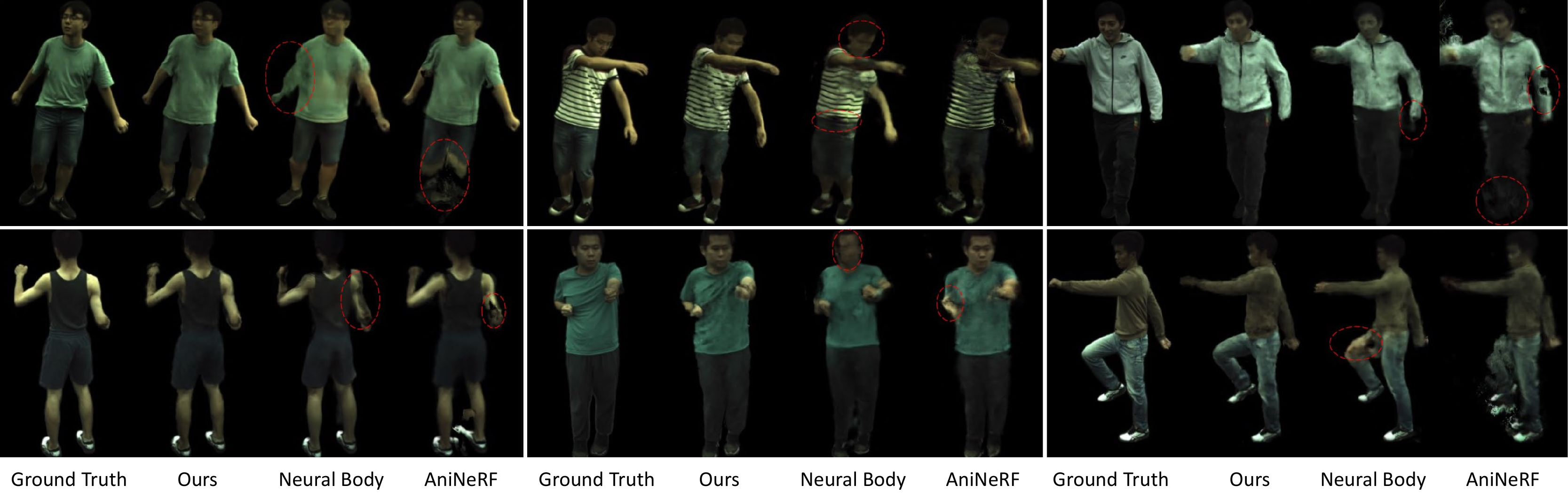}
	\caption{Results of novel pose synthesis on the ZJU-MoCap \cite{peng2021neural} dataset. Our results have fewer artifacts and are more visually pleasing. Note how our results better preserve the details on the faces of the subjects.}
	\label{fig:novel_pose_zju}
\end{figure*}

\begin{figure*}[t]
	\centering
	\includegraphics[width=1\linewidth]{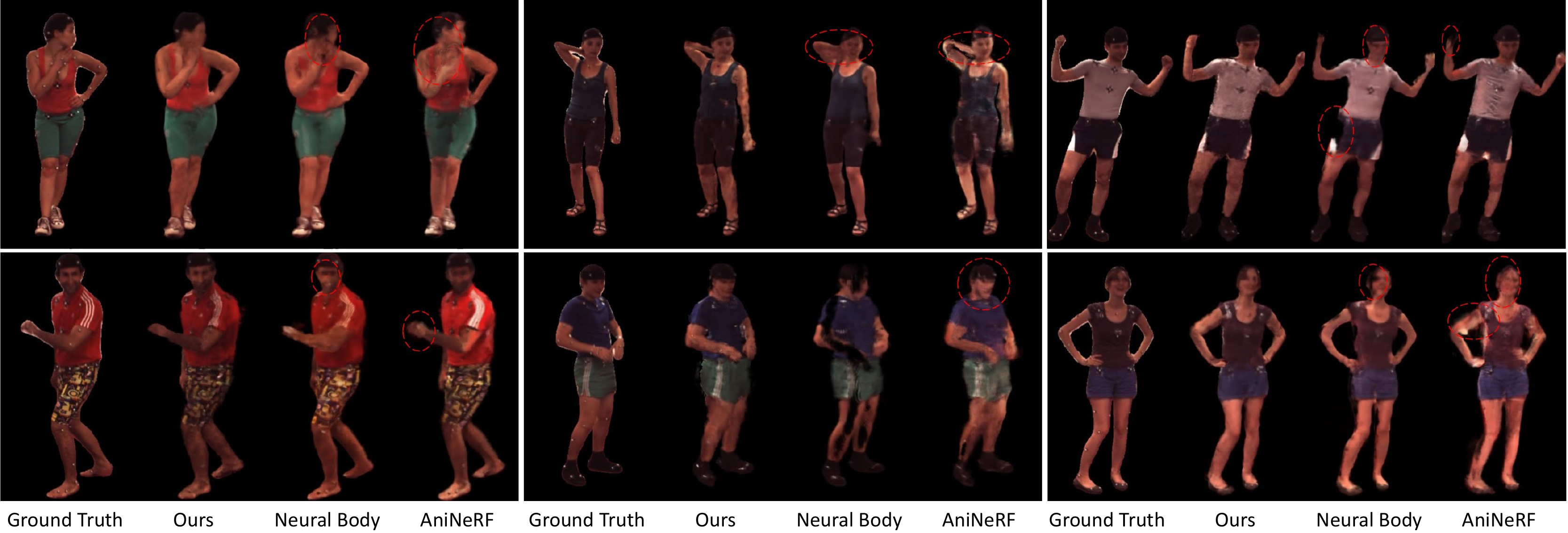}
	\caption{Results of novel pose synthesis on the Human3.6M \cite{ionescu2013human3} dataset. Our results show higher visual quality with fewer artifacts. Also, note the realistic lighting and shading effects in our results.}
	\label{fig:novel_pose_h36m}
\end{figure*}

\textbf{Datasets.} ZJU-MoCap \cite{peng2021neural} is a multi-view dataset containing 9 performers captured by 21 synchronized cameras. It provides estimated SMPL \cite{SMPL:2015} parameters and instance-level human-parsing masks generated by an established method \cite{gong2018instance}. We follow the experimental settings of Neural Body \cite{peng2021neural} and AniNeRF \cite{peng2021animatable}. Images corresponding to four uniformly distributed cameras are used for training and the rest for evaluation. We conduct experiments on 8 performers. 
Human3.6M \cite{ionescu2013human3} contains four-view videos with human poses collected by a marker-based motion capture system. Images corresponding to three views are used for training and one for evaluation. We use the same protocol as Neural Body \cite{peng2021neural} to generate SMPL \cite{SMPL:2015} parameters and masks.

\begin{figure*}[t]
	\centering
	\includegraphics[width=1\linewidth]{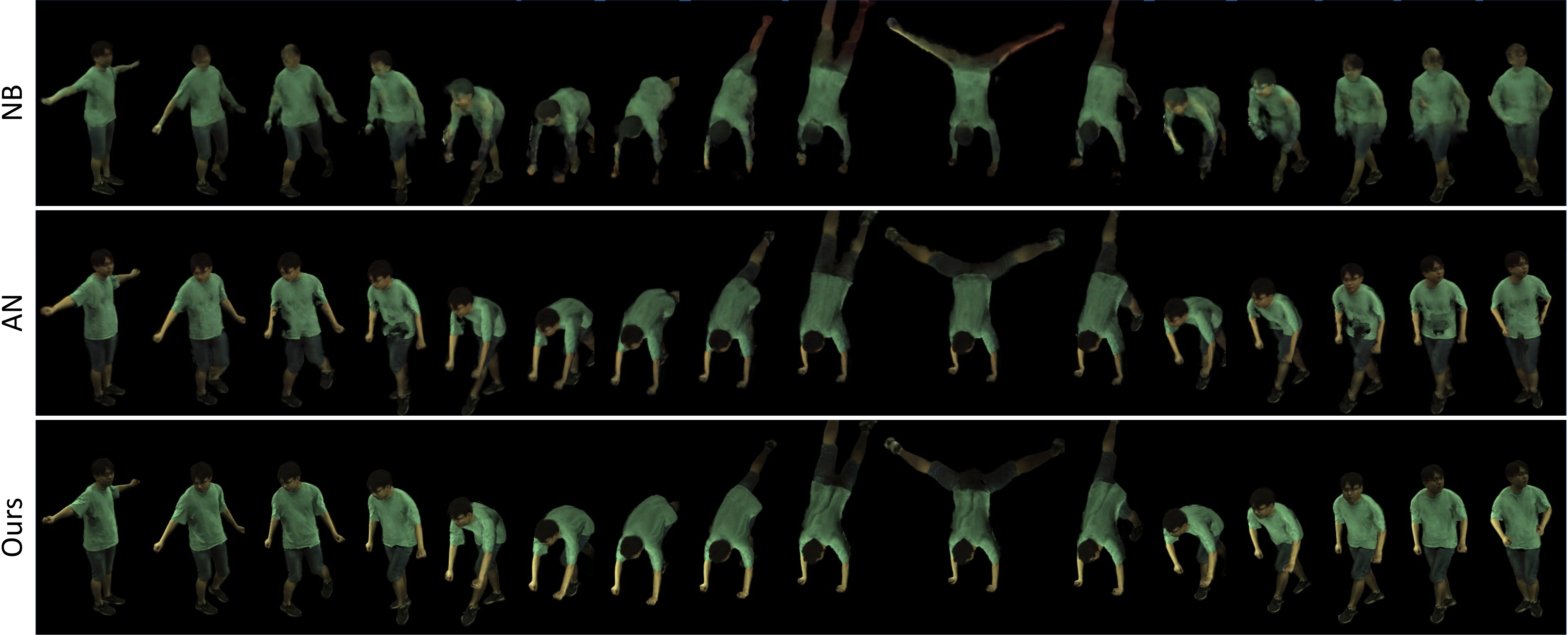}
	\caption{Results of extreme pose synthesis on the ``hand stand" sequence from AMASS \cite{AMASS:ICCV:2019}. ``NB" means Neural Body\cite{peng2021neural}, and ``AN" means AniNeRF \cite{peng2021animatable}. Neural Body produces corrupted limbs and faces. AniNeRF makes artifacts and blurs. Our method renders sharp images with clear details and realistic lighting.}
	\label{fig:extremepose}
\end{figure*}

\textbf{Metrics.} We adopt three metrics to evaluate the rendering quality, including PSNR, SSIM, and LPIPS \cite{zhang2018perceptual}. 
PSNR is a pixel-wise metric based on the mean squared error, which is sensitive to noises and random variations. 
SSIM measures the structural similarity based on luminance, contrast, and structure comparisons.
LPIPS \cite{zhang2018perceptual} measures the perceptual distance between an image pair in a deep feature space.
Since most pixels in the datasets belong to the background, we calculate PSNR and SSIM only within the 2D foreground mask, which is obtained by projecting the 3D bounding box on to the image plane.

\subsection{Image Synthesis in Novel Poses}
\textbf{Baselines.}
Since we focus on the generalization ability of the model under novel poses, we compare our method with two state-of-the-art methods \cite{peng2021neural,peng2021animatable} on novel pose synthesis. For results on novel view synthesis, please refer to our supplementary material. Note that we cannot compare with some recent works \cite{liu2021neural,xu2021h} since the official code is not released so far. Neural Body \cite{peng2021neural} provides result images on both ZJU-MoCap and Human3.6M datasets. So we directly evaluate their results with our metrics. AniNeRF~\cite{peng2021animatable} releases results only on Human3.6M. So we run the official code of AniNeRF on ZJU-MoCap and conduct the same evaluation as above. 

An implementation detail of NeuralBody and AniNeRF is that they only cast rays within the ground-truth human mask, which leaves them an advantage in comparisons. We argue that this operation is unreasonable because ground-truth masks are not always available for novel poses. Therefore, our method does not leverage the ground-truth human masks despite being at a disadvantage.
Since the Lighting MLP is undefined beyond the movement range of a subject in the training frames, we place the avatar at the mean position of the training sequence when querying the Lighting MLP for novel poses. 
Likewise, the latent embedding is also not defined for novel poses, so we set it to zeros as done by previous work \cite{zheng2022structured}.

\textbf{Comparisons on novel pose synthesis.} As shown in \cref{tab:zju_novel_pose} and \cref{tab:h36m_novel_pose}, our method achieves the best PSNR and SSIM scores compared to two strong baselines. Since previous works \cite{zhang2018unreasonable,liu2019liquid,liu2021neural} show that higher PSNR and SSIM scores do not guarantee better visual quality of images, we report LPIPS as a perceptual measurement, on which our method also shows advantages. According to \cref{fig:novel_pose_zju} and \cref{fig:novel_pose_h36m}, our method produces fewer artifacts than AniNeRF~\cite{peng2021animatable}, indicating a better correspondences across frames. Meanwhile, the lighting conditions in our results are closer to the ground truths and the details are easier to recognize thanks to the reasonable decoupling of body properties and the environmental lighting. As shown in \cref{fig:novel_pose_zju} and \cref{fig:novel_pose_h36m}, Neural Body \cite{peng2021neural} tends to produce wrong body structures on Human3.6M when an unseen pose is far from the seen ones in the training set. While, our method produces visually pleasing results under novel poses, demonstrating the robustness of the barycentric mapping and the correctness of our lighting model.

\begin{table}[]
	\centering
	\scriptsize
	\renewcommand{\arraystretch}{1.1}
	\renewcommand{\tabcolsep}{1.2mm}
	\begin{tabular}{@{}c|ccc|ccc|ccc@{}}
		\toprule
		\textbf{} & \multicolumn{3}{c|}{PSNR$\uparrow$}        & \multicolumn{3}{c|}{SSIM$\uparrow$} & \multicolumn{3}{c}{LPIPS$\downarrow$} \\ \cmidrule(l){2-10} 
		& NB              & AN     & Ours            & NB      & AN      & Ours            & NB       & AN      & Ours             \\ \midrule
		Twirl     & 23.853          & 22.800 & \textbf{24.300} & 0.902   & 0.863   & \textbf{0.910}  & 0.056    & 0.078   & \textbf{0.045}   \\
		Taichi    & \textbf{19.606} & 18.470 & 19.533          & 0.853   & 0.795   & \textbf{0.862}  & 0.054    & 0.092   & \textbf{0.047}   \\
		Warmup    & 23.907          & 23.280 & \textbf{24.675} & 0.909   & 0.901   & \textbf{0.919}  & 0.036    & 0.056   & \textbf{0.031}   \\
		Punch1    & 25.671          & 25.550 & \textbf{26.042} & 0.881   & 0.872   & \textbf{0.889}  & 0.044    & 0.053   & \textbf{0.035}   \\
		Punch2    & 21.595          & 21.916 & \textbf{22.395} & 0.870   & 0.838   & \textbf{0.881}  & 0.058    & 0.089   & \textbf{0.050}   \\
		Swing1    & 25.736          & 18.438 & \textbf{25.776} & 0.908   & 0.670   & \textbf{0.914}  & 0.049    & 0.212   & \textbf{0.045}   \\
		Swing2    & 23.802          & 21.870 & \textbf{24.360} & 0.878   & 0.836   & \textbf{0.888}  & 0.055    & 0.090   & \textbf{0.049}   \\
		Swing3    & \textbf{23.248} & 17.694 & 23.247          & 0.893   & 0.792   & \textbf{0.894}  & 0.055    & 0.206   & \textbf{0.053}   \\ \midrule
		Average   & 23.427          & 21.252 & \textbf{23.791} & 0.887   & 0.821   & \textbf{0.894}  & 0.051    & 0.110   & \textbf{0.044}   \\ \bottomrule
	\end{tabular}
	\caption{Comparison with baselines on novel pose synthesis on ZJU-MoCap \cite{peng2021neural}, ``NB" means Neural Body \cite{peng2021neural}, and ``AN" means AniNeRF \cite{peng2021animatable}. Our method outperforms both baselines with a clear margin.}
	\label{tab:zju_novel_pose}
\end{table}

\begin{table}[]
	\centering
	\scriptsize
	\renewcommand{\arraystretch}{1.1}
	\renewcommand{\tabcolsep}{1.2mm}
	\begin{tabular}{@{}c|ccc|ccc|ccc@{}}
		\toprule
		\textbf{}      &    \multicolumn{3}{c|}{PSNR$\uparrow$}     &   \multicolumn{3}{c|}{SSIM$\uparrow$}   &   \multicolumn{3}{c}{LPIPS$\downarrow$}   \\
		\cmidrule(l){2-10} &       NB        &   AN   &      Ours       &       NB       &  AN   &      Ours      &        NB        &  AN   &      Ours      \\ \midrule
		S1         &     21.932      & 19.955 & \textbf{23.206} &     0.873      & 0.855 & \textbf{0.886} & {\textbf{0.026}} & 0.029 &     0.027      \\
		S5         & \textbf{23.332} & 20.022 &     23.025      & \textbf{0.893} & 0.840 &     0.886      &     {0.022}      & 0.025 & \textbf{0.021} \\
		S6         &     23.263      & 23.637 & \textbf{24.059} &     0.888      & 0.882 & \textbf{0.893} &     {0.041}      & 0.046 & \textbf{0.038} \\
		S7         &     22.398      & 21.762 & \textbf{22.913} & \textbf{0.888} & 0.869 &     0.885      &     {0.029}      & 0.033 & \textbf{0.027} \\
		S8         &     20.779      & 21.631 & \textbf{22.659} &     0.872      & 0.877 & \textbf{0.889} &     {0.035}      & 0.032 & \textbf{0.031} \\
		S9         &     22.868      & 21.948 & \textbf{24.143} &     0.880      & 0.871 & \textbf{0.887} & {\textbf{0.029}} & 0.034 & \textbf{0.028} \\
		S11         &     23.538      & 22.547 & \textbf{24.842} &     0.879      & 0.875 & \textbf{0.894} &     {0.032}      & 0.030 & \textbf{0.029} \\ \midrule
		Average       &     22.587      & 21.643 & \textbf{23.550} &     0.882      & 0.867 & \textbf{0.889} &     {0.030}      & 0.033 & \textbf{0.029} \\ \bottomrule
	\end{tabular}
	\caption{Comparison with baselines on novel pose synthesis on the Human3.6M \cite{ionescu2013human3} dataset. ``NB" means Neural Body~\cite{peng2021neural}, and ``AN" means AniNeRF \cite{peng2021animatable}. Our method outperforms both baselines in most cases.}
	\label{tab:h36m_novel_pose}
\end{table}

\textbf{Comparison on extreme pose synthesis.}
Since the difference between the training and the test poses in a dataset may not be large enough, we compare the methods on a challenging pose sequence from the AMASS~\cite{AMASS:ICCV:2019} database. As shown in \cref{fig:extremepose}, 
Neural Body produces corrupted limbs and faces, which show the limitation of the convolution-based solution. AniNeRF makes artifacts and blurs due to the instability of the spatially interpolated LBS weights and the poor generalization ability of the neural blending weights. Our method renders sharp images with clear details and realistic lighting thanks to the stable correspondences and reasonable decoupling of body properties and the environmental lighting. Please refer to our supplementary video for animated results.

\subsection{Ablation Studies}
To verify the effectiveness of our main components, we conduct ablation studies on the ``Twirl" sequence of ZJU-MoCap \cite{peng2021neural} in terms of novel pose and novel view synthesis. All models are trained for the same number of epochs (100) for a fair comparison.

\begin{figure}[h!]
	\centering
	\includegraphics[width=1.0\linewidth]{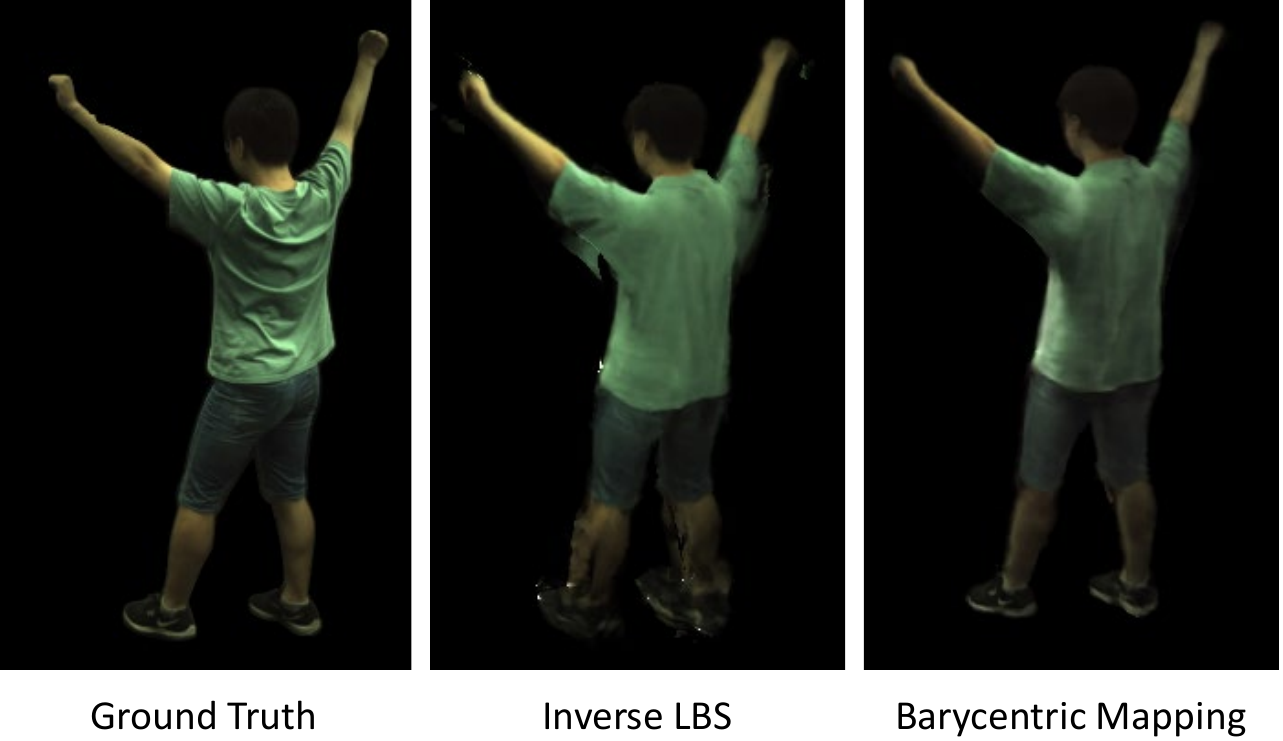}
	\caption{Visual comparison between the barycentric mapping and inverse LBS. Inverse LBS with interpolated blending weights produces artifacts near movement-frequent places like armpits and feet while the barycentric mapping renders clear results.
	}
	\label{fig:ablation_mapping}
\end{figure}

\begin{figure}[h!]
	\centering
	\includegraphics[width=1.0\linewidth]{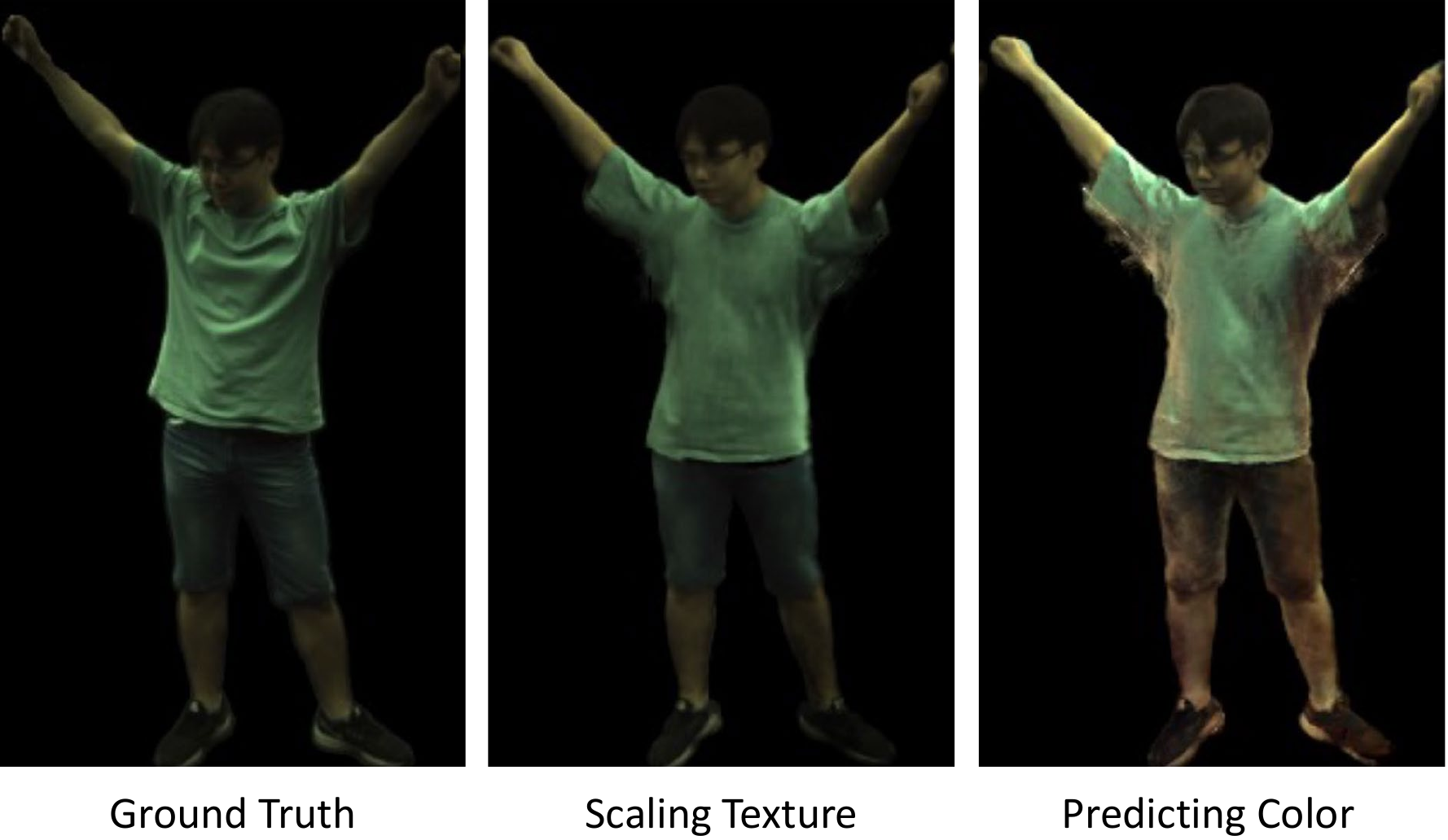}
	\caption{Ablation study of the Lighting MLP. We test with an alternative design of Lighting MLP that directly predicts RGB values instead of predicting the lightness coefficient for texture scaling. However, the Lighting MLP that directly predicts colors tends to overfit the training frames and produces distorted colors on the novel pose.}
	\label{fig:ablation_lightingmlp_design}
\end{figure}

\begin{table}[h!]
	\centering
	\scriptsize
	\renewcommand{\arraystretch}{1.1}
	\renewcommand{\tabcolsep}{1.3mm}
	\begin{tabular}{@{}c|cc|cc|cc@{}}
		\toprule
		\textbf{}                   & \multicolumn{2}{c|}{PSNR$\uparrow$}                      & \multicolumn{2}{c|}{SSIM$\uparrow$} & \multicolumn{2}{c}{LPIPS$\downarrow$} \\ \cmidrule(l){2-7} 
		& View                       & Pose                        & View             & Pose             & View              & Pose              \\ \midrule
		\textbf{Full model}         & \underline{31.090}               & \textbf{24.216}             & \underline{0.970}      & \textbf{0.911}   & \underline{0.023}       & \textbf{0.044}    \\
		Replace BM with inverse LBS & 30.758                     & 23.301                      & 0.968            & 0.895            & 0.026             & 0.055             \\
		w/o Lighting MLP            & \multicolumn{1}{l}{30.696} & \multicolumn{1}{l|}{23.465} & 0.967            & \underline{0.906}      & 0.027             & 0.049             \\
		Lighting MLP (predicting color) & \textbf{31.270}            & \underline{23.570}                & \textbf{0.971}   & 0.905            & \textbf{0.019}    & \underline{0.047}       \\
		\bottomrule
	\end{tabular}
	\caption{Ablation studies. ``View" refers to novel view synthesis, and ``Pose" refers to novel pose synthesis. ``BM" means the barycentric mapping. Bold values are the best scores, and underlined values are the second best.}
	\label{tab:ablation}
\end{table}

\textbf{Barycentric mapping.} 
To validate the virtue of the barycentric mapping, we replace it with the inverse LBS algorithm. For the blending weights, we follow the strategy of AniNeRF \cite{peng2021animatable}, which interpolates the blending weights from nearby SMPL vertices. The second row of \cref{tab:ablation} shows the clear superiority of the barycentric mapping, especially on novel poses. In the visual comparison (\cref{fig:ablation_mapping}), inverse LBS produces artifacts around movement-frequent places like armpits and feet, while the barycentric mapping still performs well.

\textbf{w/o Lighting MLP.} 
The Lighting MLP plays a vital role in solving the lighting inconsistency. It models the correct location-dependent lighting in the world space and benefits high-fidelity rendering. To validate its usefulness, we disable the Lighting MLP when training and rendering. Then, our model degenerates to a vanilla NeRF defined in the canonical space. The third row in \cref{tab:ablation} shows significant degradation in all metrics.

\textbf{Lighting MLP predicts color directly.}
We also try an alternative design of the Lighting MLP that takes in the body texture and outputs the color instead of predicting the lightness coefficient for texture scaling. As shown in the fourth row in \cref{tab:ablation}, this alternative Lighting MLP perform better on novel view synthesis due to higher expressiveness but perform worse on novel poses. Thus, the alternative Lighting MLP is just overfitting the colors in the training frames instead of learning the environmental lighting. Similar conclusion can be drawn from \cref{fig:ablation_lightingmlp_design}, where the alternative Lighting MLP predicts distorted colors on the novel pose. We show animated results in our supplementary video.

\section{Conclusion}
In this paper, we focus on the generalization problem of human body reconstruction and animation. We propose to model the human body and the lighting condition in separate spaces. To bridge the canonical space and the world space, we propose the barycentric mapping, which helps us to transform point positions and surface normals of a human body between the two spaces, enabling rendering in the world space with body properties from the canonical space. Most importantly, the barycentric mapping can directly generalize to novel poses without additional input or network training. Thanks to the reasonable decoupling of body properties and lighting conditions, we obtain clear improvements upon two strong baselines.

\section{Limitations and Potential Impacts}
Our method uses SMPL \cite{SMPL:2015} as a proxy to build connections between the world space and the canonical space. Therefore, it strongly relies on an accurate SMPL fitting. In scenarios where SMPL parameters cannot be precisely obtained, our method is likely to fail. Also, our approach does not model long-range dependencies and thus is unable to deal with a performer in a long dress. Our work reconstructs the appearances of subjects and animates them with public video datasets. Currently, the rendering realism is far from fooling people, but attention should be paid to future versions of related technologies for potential misusing.

\noindent\textbf{Acknowledgments:}
The work is supported by National Key R\&D Program of China (2018AAA0100704), NSFC \#61932020, \#62172279, Science and Technology Commission of Shanghai Municipality (Grant No. 20ZR1436000), Program of Shanghai Academic Research Leader, and ``Shuguang Program'' supported by Shanghai Education Development Foundation and Shanghai Municipal Education Commission.

{\small
	\bibliographystyle{ieee_fullname}
	\bibliography{egbib}
}

\clearpage

\section{Implementation Details}
Empirically, the points that are too far away from the SMPL \cite{SMPL:2015} mesh should not follow the movement of the human body. To make the learned neural radiance field less noisy, we exclude these outlier points during volumetric rendering by setting their density values to zeros, and this operation brings a slight improvement. To determine whether a given point is an outlier, we apply the following algorithm:

\begin{algorithm}[h] 
	\caption{Outlier detection} 
	\begin{algorithmic}[1] 
		\Require 
	    a sampled point $p$; 
		SMPL faces $\mathcal{F}$; 
		$\alpha$; $\beta$; $\gamma$
		\Ensure outlier mask $m$
		\State $f \leftarrow$ Find\_NN\_Mesh($p$, $\mathcal{F}$)
		\State $u, v, h \leftarrow$ Compute\_UV\_SignedDistance($p, f$)
		\State return $(u,v< \alpha)$ OR $(u,v > \beta)$ OR $(|h| > \gamma)$
	\end{algorithmic} 
\end{algorithm}

\noindent We set $\alpha = -4$, $\beta = 5$, and $\gamma = 0.1$ experimentally.

\section{Novel View Synthesis}
\label{sec: noveview}
\begin{figure*}[h]
 \centering
  \includegraphics[width=1\linewidth]{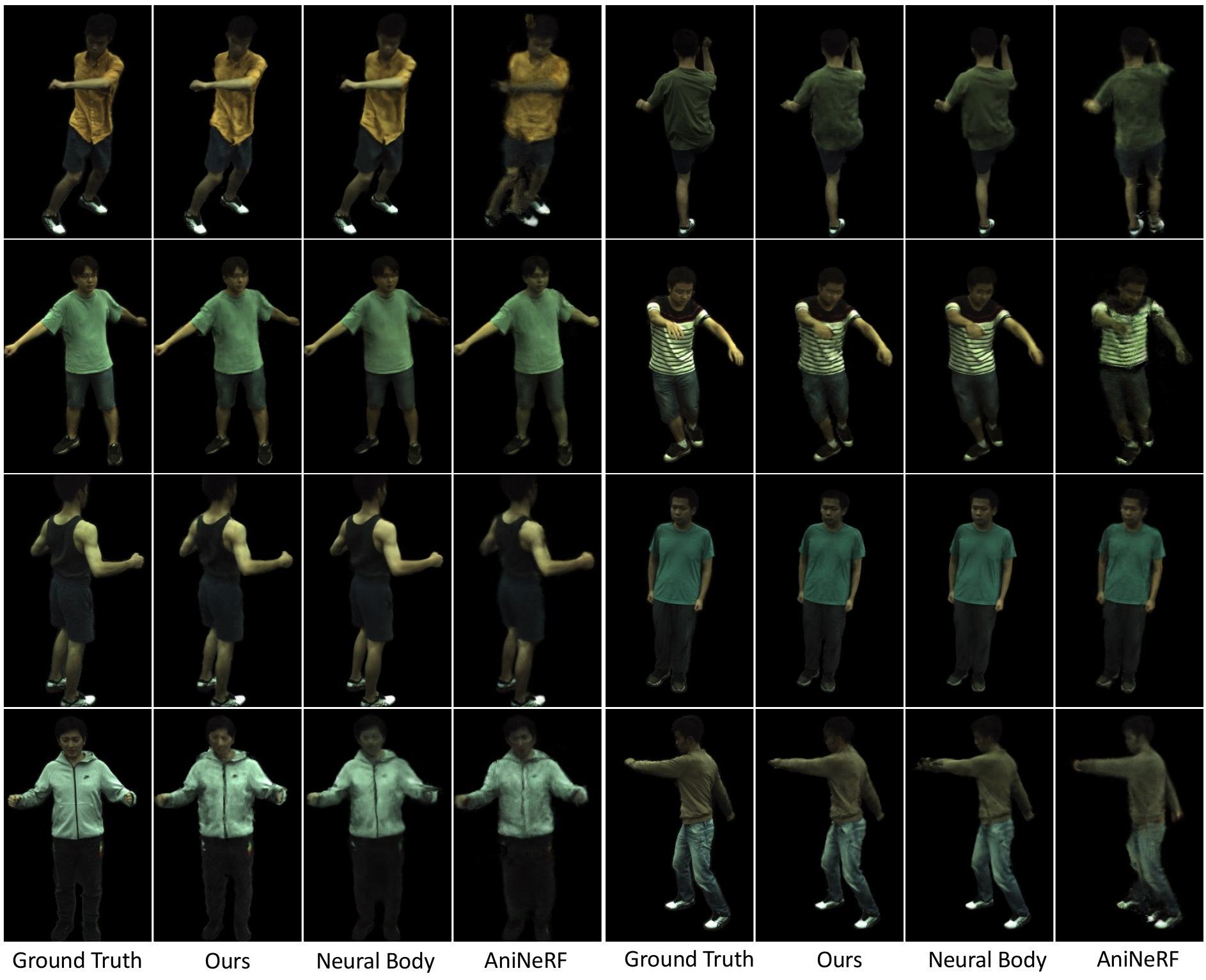}
  \caption{Results of novel view synthesis on the ZJU-MoCap \cite{peng2021neural} dataset. The results of Neural Body \cite{peng2021neural} and our method exhibit fewer artifacts compared with AniNeRF \cite{peng2021animatable}.}
  \label{fig:novel_view_zju}
\end{figure*}

\begin{figure*}[h]
 \centering
  \includegraphics[width=1\linewidth]{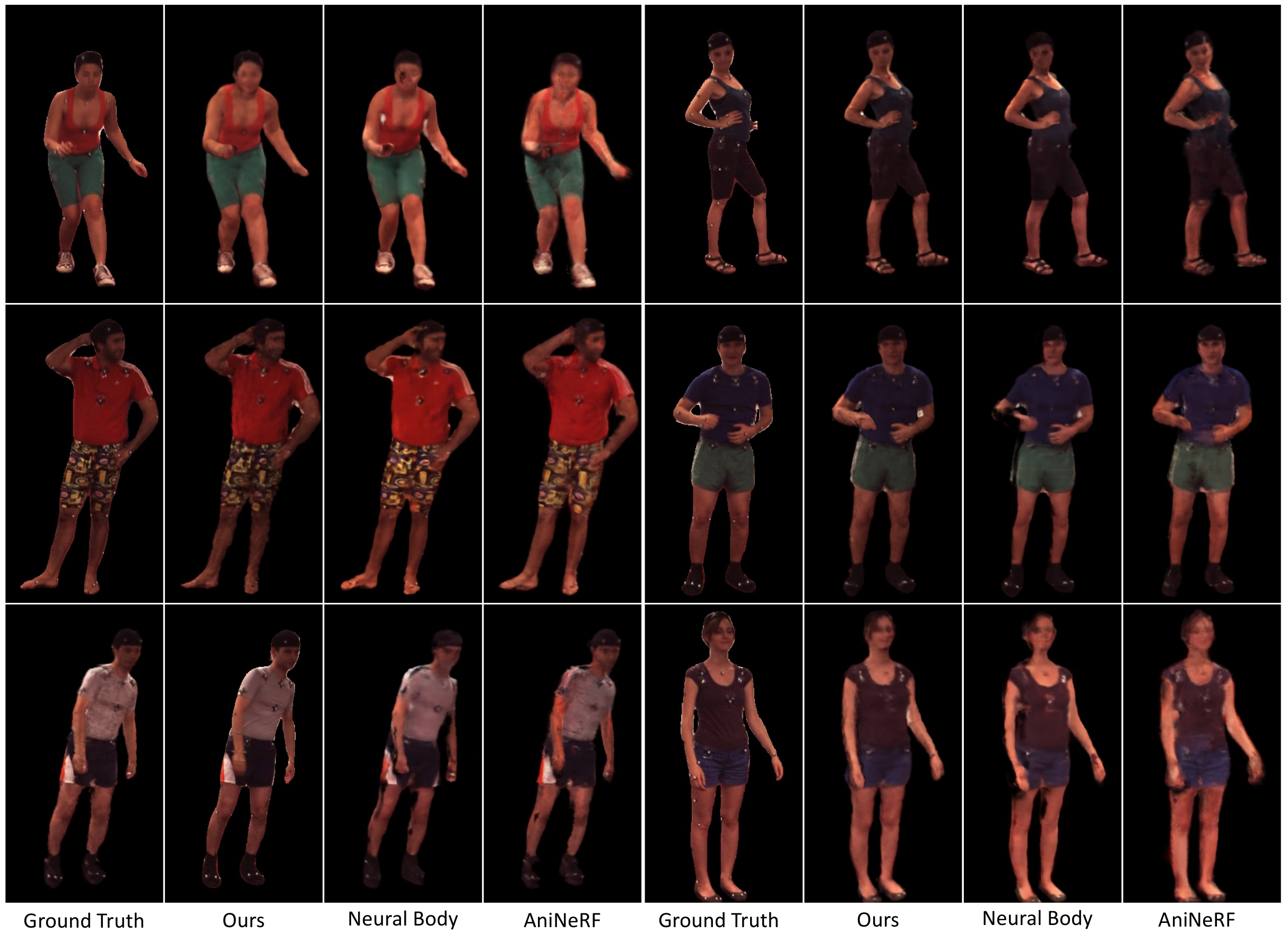}
  \caption{Results of novel view synthesis on the Human3.6M \cite{ionescu2013human3} dataset. Our results show higher fidelity with clear textures and realistic lighting and shading.}
  \label{fig:novel_view_h36m}
\end{figure*}

Although our method targets novel pose synthesis, we still report results on novel view synthesis. We compare the methods on the ZJU-Mocap \cite{peng2021neural} dataset in \cref{tab:zju_novel_view}, where Neural Body \cite{peng2021neural} exhibits superiority on PSNR and SSIM, and achieves comparable LPIPS to our method. Neural Body anchors features on the vertices of SMPL and diffuses them into a feature grid before volumetric rendering, avoiding establishing correspondences across frames. It is favorable for novel view synthesis but degrades on novel poses. Our method exceeds AniNeRF \cite{peng2021animatable}, which explicitly builds correspondences across frames like our method, by a large margin in all metrics. And the results of quantitative comparisons are consistent with the rendering results shown in \cref{fig:novel_view_zju}. Our barycentric mapping establishes more robust correspondences across poses, producing fewer structural artifacts such as the extra feet in \cref{fig:novel_view_zju}.
On the Human3.6M \cite{ionescu2013human3} dataset, our method shows outstanding performance compared to both baselines as shown in \cref{tab:h36m_novel_view}. Corresponding visual comparisons are shown in \cref{fig:novel_view_h36m}, where our method produces realistic textures, lights, and shades. Note that Human3.6M \cite{ionescu2013human3} is noisier with higher errors in the fitted SMPL parameters and unclear boundaries in foreground masks compared to ZJU-MoCap \cite{peng2021neural}. This explains the obvious degradation of Neural Body on this dataset and indicates higher robustness towards imperfect SMPL fits and noisy data of our method. 

\begin{table}[t]
	\centering
	\scriptsize
	\renewcommand{\tabcolsep}{1.1mm}
    \begin{tabular}{@{}c|ccc|ccc|ccc@{}}
        \toprule
        \textbf{} & \multicolumn{3}{c|}{PSNR$\uparrow$}        & \multicolumn{3}{c|}{SSIM$\uparrow$}     & \multicolumn{3}{c}{LPIPS$\downarrow$}   \\ \cmidrule(l){2-10} 
                  & NB              & AN     & ours            & NB             & AN    & ours           & NB             & AN    & ours           \\ \midrule
        Twirl     & \underline{30.455}    & 28.050 & \textbf{31.623} & \underline{0.966}    & 0.940 & \textbf{0.972} & \underline{0.028}    & 0.038 & \textbf{0.024} \\
        Taichi    & \textbf{27.199} & 19.660 & \underline{25.829}    & \textbf{0.960} & 0.849 & \underline{0.950}    & \textbf{0.022} & 0.065 & \underline{0.026}    \\
        Warmup    & \textbf{27.962} & 24.970 & \underline{27.347}    & \textbf{0.952} & 0.920 & \underline{0.950}    & \textbf{0.026} & 0.045 & \textbf{0.026} \\
        Punch1    & \textbf{28.659} & 25.760 & \underline{28.487}    & \textbf{0.928} & 0.870 & \underline{0.925}    & \underline{0.027}    & 0.054 & \textbf{0.026} \\
        Punch2    & \textbf{25.866} & 22.551 & \underline{25.208}    & \textbf{0.927} & 0.862 & \underline{0.918}    & \underline{0.045}    & 0.083 & \textbf{0.044} \\
        Swing1    & \textbf{29.618} & 23.717 & \underline{29.226}    & \textbf{0.946} & 0.869 & \underline{0.942}    & \textbf{0.030} & 0.074 & \underline{0.031}    \\
        Swing2    & \textbf{28.632} & 23.793 & \underline{28.494}    & \textbf{0.939} & 0.877 & \underline{0.933}    & \textbf{0.034} & 0.069 & \underline{0.035}    \\
        Swing3    & \textbf{27.583} & 17.351 & \underline{27.199}    & \textbf{0.936} & 0.760 & \underline{0.930}    & \textbf{0.034} & 0.205 & \underline{0.035}    \\ \midrule
        Average   & \textbf{28.247} & 23.232 & \underline{27.927}    & \textbf{0.944} & 0.868 & \underline{0.940}    & \textbf{0.031} & 0.079 & \textbf{0.031} \\ \bottomrule
    \end{tabular}
    \caption{Comparison with baselines on novel view synthesis on the ZJU-MoCap \cite{peng2021neural} dataset. ``NB" means Neural Body, and ``AN" means AniNeRF \cite{peng2021animatable}. Bold values are the best scores, and underlined values are the second best. Our method outperforms AniNeRF \cite{peng2021animatable} and is comparable to Neural Body  \cite{peng2021neural} on the perceptual metric.}
    \label{tab:zju_novel_view}
\end{table}

\begin{table}[t]
	\centering
    \scriptsize
	\renewcommand{\tabcolsep}{0.9 mm}
    \begin{tabular}{@{}c|ccc|ccc|ccc@{}}
        \toprule
        \textbf{} & \multicolumn{3}{c|}{PSNR$\uparrow$}           & \multicolumn{3}{c|}{SSIM$\uparrow$}           & \multicolumn{3}{c}{LPIPS$\downarrow$}         \\ \cmidrule(l){2-10} 
                  & NB           & AN           & ours            & NB             & AN          & ours           & NB             & AN          & ours           \\ \midrule
        S1        & \underline{22.716} & 22.415       & \textbf{24.494} & \underline{0.893}    & 0.890       & \textbf{0.913} & \textbf{0.032} & 0.034       & \textbf{0.032} \\
        S5        & \underline{24.439} & 23.228       & \textbf{24.819} & \underline{0.914}    & 0.891       & \textbf{0.915} & \underline{0.023}    & 0.027       & \textbf{0.022} \\
        S6        & 22.668       & \underline{22.689} & \textbf{24.294} & \underline{0.884}    & 0.866       & \textbf{0.894} & \underline{0.029}    & 0.034       & \textbf{0.028} \\
        S7        & \underline{22.991} & 21.793       & \textbf{23.933} & \textbf{0.911} & 0.886       & \underline{0.909}    & \textbf{0.025} & 0.030       & \underline{0.026}    \\
        S8        & 21.570       & \underline{22.666} & \textbf{23.234} & 0.890          & \underline{0.897} & \textbf{0.911} & 0.034          & \underline{0.031} & \textbf{0.027} \\
        S9        & 24.121       & \underline{24.694} & \textbf{25.691} & \underline{0.907}    & \underline{0.907} & \textbf{0.914} & \textbf{0.029} & 0.034       & \textbf{0.029} \\
        S11       & 23.537       & \underline{24.594} & \textbf{25.622} & 0.892          & \underline{0.903} & \textbf{0.914} & 0.039          & \underline{0.035} & \textbf{0.032} \\ \midrule
        Average   & 23.149       & \underline{23.154} & \textbf{24.584} & \underline{0.899}    & 0.891       & \textbf{0.910} & \underline{0.030}    & 0.032       & \textbf{0.028} \\ \bottomrule
    \end{tabular}
    \caption{Comparison with baselines on novel view synthesis on the Human3.6M \cite{ionescu2013human3} dataset. ``NB" means Neural Body \cite{peng2021neural}, and ``AN" means AniNeRF \cite{peng2021animatable}. Bold values are the best scores, and underlined values are the second best. Our method achieves the highest performances on PSNR, SSIM, and LPIPS. }
    \label{tab:h36m_novel_view}
\end{table}

\section{Novel Pose Synthesis}
To further verify the generalization ability of our method on novel poses, we show animated results in our supplementary video. We animate the subjects of each dataset with a pose sequence from the other dataset. In pursuit of diversity and complexity, we select the ``Swing3" sequence from ZJU-Mocap and the ``S9" sequence from Human3.6M. Besides, we compare our method with the baselines on three more challenging pose sequences from the AMASS \cite{AMASS:ICCV:2019} database. Neural Body \cite{peng2021neural} can hardly generalize to extreme poses. AniNeRF \cite{peng2021animatable} lacks high-fidelity details such as wrinkles and lighting. In extreme pose \#3, AniNeRF produces artifacts like corrupted faces, while our method gives stable results.

\section{Lighting MLP Validation}
To further interpret our Lighting MLP, we visualize its effect in our supplementary video by manipulating the querying points of the Lighting MLP. By rotating the querying points along a certain axis, we can observe a change in the lighting on the human body. While, if the subject walks out of the moving boundary in the training frames, the rendering results will be dim because those locations are undefined for the Lighting MLP.


\end{document}